\documentclass[11pt]{article}
\usepackage{coling2020}
\usepackage{times}
\usepackage{url}
\usepackage{latexsym}
\usepackage{hyperref}
\usepackage{amsmath, amssymb}
\usepackage{multirow}
\usepackage{bbm}

\usepackage{color}
\usepackage{caption}
\usepackage[labelformat=simple]{subcaption}

\usepackage{mathrsfs}
\usepackage{booktabs}
\usepackage{graphicx, float}

\usepackage[normalem]{ulem}

\usepackage{microtype}

\DeclareMathOperator*{\softmax}{softmax}
\DeclareMathOperator*{\bilstm}{BiLSTM}
\DeclareMathOperator*{\argmax}{arg\,max}
\usepackage[table, dvipsnames]{xcolor}
\colingfinalcopy 

\title{Exploring End-to-End Differentiable Natural Logic Modeling}
\author{
Yufei Feng$^\dagger$\footnotemark[1],
~~Zi'ou Zheng$^\dagger$\footnotemark[1],
~~Quan Liu$^\ddagger$,  
~~\textbf{Michael Greenspan}$^\dagger$, 
~~\textbf{Xiaodan Zhu}$^\dagger$ \\ 
  $^\dagger$Electrical and Computer Engineering
  \& Ingenuity Labs Research Institute, Queen's University \\ 
  $^\ddagger$State Key Laboratory of Cognitive Intelligence, iFLYTEK Research \\
  
  \texttt{\{feng.yufei,ziou.zheng,michael.greenspan,xiaodan.zhu\}@queensu.ca} \quad \\ 
  \texttt{quanliu@iflytek.com}
  }

\date{}

\begin{document}
\maketitle
\renewcommand{\thefootnote}{\fnsymbol{footnote}} 
\footnotetext[1]{Equal contribution.}
\renewcommand{\thefootnote}{\arabic{footnote}}
\begin{abstract}
We explore end-to-end trained differentiable models that integrate natural logic with neural networks, aiming to keep the backbone of natural language reasoning based on the natural logic formalism while introducing subsymbolic vector representations and neural components. The proposed model adapts module networks to model natural logic operations, which is enhanced with a memory component to model contextual information.   Experiments show that the proposed framework can effectively model monotonicity-based reasoning, compared to the baseline neural network models without built-in inductive bias for monotonicity-based reasoning. Our proposed model shows to be robust when transferred from upward to downward inference. We perform further analyses on the performance of the proposed model on aggregation, showing the effectiveness of the proposed subcomponents on helping achieve better intermediate aggregation performance. 


\end{abstract}

\section{Introduction}
\blfootnote{

    %
    \hspace{-0.50cm}  
    This work is licensed under a Creative Commons 
    Attribution 4.0 International License.
    License details:
    \url{http://creativecommons.org/licenses/by/4.0/}.
}

A recent research trend has attempted to further advance the long-standing problem of bringing together 
the complementary strengths of neural networks and symbolic models, e.g., the research performed in   \cite{garcez2015neural,yang2017differentiable,rocktaschel2017end,evans2018learning,weber2019nlprolog,de2019neuro,Mao2019NeuroSymbolic}, among others. It is known that neural models can approximate complex functions and are robust to noise and ambiguity, while symbolic models often render superior explainability and interpretability but are brittle and prone to fail in the presence of noise and uncertainty.

The majority of research efforts are based on some abstract logical forms such as the first-order logic (FOL) or its \textit{fragments}.
For natural language, obtaining such a representation is known to face many thorny challenges. Natural logic instead aims to sidestep some of the challenges by performing inferences over surface forms of text based on \emph{monotonicity} or \emph{projectivity} \cite{van1986essays,valencia1991studies,maccartney2009extend,icard2014recent}, and has been applied to tasks such as natural language inference~\cite{maccartney2009extend,angeli2014naturalli} and question answering~\cite{angeli2016combining}.

In this work we explore differentiable natural logic models that integrate natural logic with neural networks, with the aim to keep the backbone of inference based on the natural logic formalism, while introducing subsymbolic vector representations and neural components into the framework. Combining the advantages of neural networks with natural logic needs to take several basic problems into consideration. Two problems flow directly from this objective: 1) How (and where) to leverage the strength of neural networks  in the natural logic formalism, and; 2) How to alleviate the issue of a lack of intermediate supervision for training sub-components, which may lead to the spurious problem~\cite{guu2017language,min2019discrete} in the end-to-end training.

We explore a framework in which module networks \cite{Andreas_2016_CVPR,gupta2019neural} are leveraged to model the natural logic operations, which is enhanced with a memory module component to capture contextual information. At the lexical and local relation learning layers, we constrain the network to predict the seven natural logic relations. The entire model is differentiable and end-to-end trained.  




We evaluate and analyze the proposed model on the monotonicity subset of Semantic Fragments~\cite{fragments2020}, HELP~\cite{help2019} and MED~\cite{med2019}. We also extend MED to generate a dataset to help evaluate 2-hop inference. The model can effectively learn natural logic operations in the end-to-end training paradigm.\footnote{Our code is available at  \url{https://github.com/feng-yufei/Neural-Natural-Logic}}

\begin{table}
\centering
\setlength{\tabcolsep}{5pt}
\begin{tabular}{|cccc|}

\hline
\textbf{Relation} & \textbf{Relation Name} & \textbf{Example} & \textbf{Set Theoretic Definition}  \\
\hline
  $x \equiv y$  & equivalence &  $mom \equiv mother$  & $x = y$   \\
  $x \sqsubset y$  & forward entailment &  $cat \sqsubset animal$ &  $x \subset y$ \\
  $x \sqsupset y$  & reverse entailment &  $animal \sqsupset cat$  &  $x \supset y$   \\
  $x$ \textsuperscript{$\wedge$} $y$  & negation & $human$ \textsuperscript{$\wedge$} $ nonhuman$   & $x \cap y = \varnothing \wedge x \cup y = U$  \\
  $x \mid y$  & alternation & $cat \mid dog$   & $x \cap y = \varnothing \wedge x \cup y \ne U$  \\
  $x \smallsmile y$  & cover & $ animal  \smallsmile nonhuman $   & $x \cap y \ne  \varnothing  \wedge x \cup y = U$   \\
  $x \ \# \ y$  & independence & $ happy \   \#  \ student$   & all other cases   \\
\hline

\end{tabular}
\caption{Seven natural logic relations proposed by~\protect\newcite{maccartney2009extend}.} 

\label{table:basic-rel}
\end{table}

\definecolor{lightgray}{gray}{0.90}
\begin{table}
\begin{minipage}[!t]{0.48\columnwidth}
  \centering
  \setlength{\tabcolsep}{5pt}\small
\begin{tabular}{|c|c|ccccccc|}

\hline

\rowcolor{lightgray}[5pt][5pt]
&  & \multicolumn{7}{c|}{\textbf{Input Relation $r$}}\\ 

\rowcolor{lightgray}[5pt][5pt]
\multirow{-2}{*}{\textbf{Quantifier}} & \multirow{-2}{*}{\textbf{Projection}}& $\equiv$ &  $\sqsubset$ &   $\sqsupset$ &   $\wedge$  & $\mid$    &   $\smallsmile$  &    $\#$  \\
\hline
\multirow{2}{*}{\textit{all}} & $\rho^{arg1}(r)$ & $\equiv$  &   $\sqsupset$ &   $\sqsubset$ &   $\mid$  & $\#$    &   $\mid$   &    $\#$  \\
 & $\rho^{arg2}(r)$ & $\equiv$  &   $\sqsubset$ &   $\sqsupset$ &   $\mid$  & $\mid$    &   $\#$   &    $\#$  \\
\hline
\multirow{2}{*}{\textit{some}} & $\rho^{arg1}(r)$ & $\equiv$  &   $\sqsubset$ &   $\sqsupset$ &   $\smallsmile$  & $\#$    &   $\smallsmile$   &    $\#$  \\
 & $\rho^{arg2}(r)$ & $\equiv$  &   $\sqsubset$ &   $\sqsupset$ &   $\smallsmile$  & $\#$    &   $\smallsmile$   &    $\#$  \\
\hline
\multirow{2}{*}{\textit{no}} & $\rho^{arg1}(r)$ & $\equiv$  &   $\sqsupset$ &   $\sqsubset$ &   $\mid$  & $\#$    &   $\mid$   &    $\#$  \\
 & $\rho^{arg2}(r)$ & $\equiv$  &   $\sqsupset$ &   $\sqsubset$ &   $\mid$  & $\#$    &   $\mid$   &    $\#$  \\
\hline


\end{tabular}
\caption{The projection function $\rho$ can project an input relation $r$ into a different relation depending on the context. Here we show the projection function for each argument position for quantifier \textit{all}, \textit{some} and \textit{no}.
}\label{table:all}
  \end{minipage}
\hfil
\begin{minipage}[!t]{0.48\columnwidth}
  \centering
\begin{tabular}{|>{\columncolor{lightgray}}c|c|c|c|c|c|c|c|}
\hline
\rowcolor{lightgray}
$\Join$ & $\equiv$  &   $\sqsubset$ &   $\sqsupset$ &   $\wedge$  & $\mid$    &   $\smallsmile$   &    $\#$  \\
\hline
$\equiv$ & $\equiv$  &   $\sqsubset$ &   $\sqsupset$ &   $\wedge$  & $\mid$    &   $\smallsmile$   &    $\#$  \\
$\sqsubset$ & $\sqsubset$  &   $\sqsubset$ &   $\#$ &   $\mid$  & $\mid$    &   $\#$   &    $\#$  \\

$\sqsupset$ & $\sqsupset$  &   $\#$ &   $\sqsupset$ &   $\smallsmile$  & $\#$    &   $\smallsmile$   &    $\#$  \\

$\wedge$ &  $\wedge$  &   $\smallsmile$ &   $\mid$ &   $\equiv$  & $\sqsupset$    &   $\sqsubset$   &    $\#$  \\

$\mid$ &  $\mid$  &   $\#$ &   $\mid$ &   $\sqsubset$  & $\#$    &   $\sqsubset$   &    $\#$  \\

$\smallsmile$ & $\smallsmile$  &   $\smallsmile$ &   $\#$ &  $\sqsupset$ &   $\sqsupset$  & $\#$    &    $\#$    \\

$\#$ &   $\#$&   $\#$&   $\#$&   $\#$&   $\#$&   $\#$&   $\#$ \\
\hline
\end{tabular}
\caption{Relation aggregation table~\protect\cite{icard2012inclusion}. Relations listed in the first column are aggregated with those listed in the first row, yielding the relations in the corresponding entries in the table.}
\label{table:union}
\end{minipage}

\end{table}


\section{Related Work}
\subsection{Neural Symbolic Models}
A growing number of research efforts have recently revisited the long-standing problem of bringing together the complementary advantages of neural networks and symbolic methods.
There are at least two approaches that have received intensive attention. One uses symbolic constraints as regularizers to equip neural models with the corresponding inductive bias  \cite{demeester2016lifted,diligenti2017semantic,donadello2017logic,xu2018semantic,li2019augmenting}. 
Another approach develops differentiable end-to-end trained frameworks based on symbolic models. For example, the work in \cite{rocktaschel2017end,weber2019nlprolog,minervini2019differentiable} proposes a differentiable backward-chaining algorithm, and \newcite{dong2019neural} adopt probabilistic tensor representations for logic predicates and mimic the forward-chaining proof. \newcite{evans2018learning} treat inductive logic programming as a satisfiability problem and \newcite{deepproblog} combine high-level symbolic oriented reasoning with low-level neural perception models. The second approach is more interesting to us for exploring powerful reasoning models with built-in explainability. Unlike the existing work based on abstract logical forms, this paper explores the integration of neural networks with natural logic.

\subsection{Natural Logic} 
Natural logic \cite{lakoff1970linguistics,van1988semantics,valencia1991studies,van1995language,nairn2006computing,maccartney2009natural,icard2012inclusion,angeli2016combining} has a long history that is traceable to the syllogisms of Aristotle. It aims to model a subset of logical inferences by operating directly on the surface form and structure of language, based on monotonicity or projectivity  ~\cite{van1986essays,valencia1991studies,maccartney2009extend,icard2014recent}, rather than deduction on the abstract forms such as the first-order logic (FOL) or its fragments---it is well known that deriving logic forms for natural language is a very challenging task. 

In natural language processing, the framework proposed in \cite{maccartney2008modeling,maccartney2009extend} extends monotonicity-based models \cite{van1988semantics,valencia1991studies} to incorporate semantic exclusion and unifies them to consider implicatives~\cite{nairn2006computing}, which is a state-of-the-art natural logic formalism that has been used for multiple NLP tasks \cite{maccartney2009natural,angeli2014naturalli}. In this work we explore neural natural logic based on this formalism. We will briefly review the background in Section~\ref{sec:background}.

\subsection{Natural Language Inference} 
Previous work often studies natural logic in natural language inference (NLI). NLI~\cite{DBLP:conf/mlcw/DaganGM05,Iftene:W07-1421,maccartney2008modeling,maccartney2009extend,maccartney2009natural,angeli2014naturalli,snli2015}, also known as recognizing textual entailment (RTE), aims to model the logical relationships between two sentences, e.g.,\,as a binary (\textit{entailment} vs. \textit{non-entailment}) or three-way classification (\textit{entailment}, \textit{contradiction}, and \textit{neutral}). Recently deep learning algorithms have been proposed~\cite{snli2015,kim2017,esim2017,chen2017recurrent,chen2018enhancing,elmo2018,yoon2018dynamic,kiela2018dynamic,talman2018natural,yang2019enhancing,bert2019}. In this paper we will describe and evaluate our neural natural logic models on NLI. The proposed model may also be further extended to other tasks in which natural logic has been applied, e.g.,\,question answering~\cite{angeli2016combining}.

\section{Background} 
\label{sec:background}
This section briefly reviews the natural logic formalism \cite{maccartney2009extend} that our work is based on. For more details, we refer readers to ~\cite{maccartney2008modeling,maccartney2009extend,maccartney2009natural,angeli2016combining}.



Monotonicity is a pervasive feature of natural language and an essential concept in natural logic~\cite{van1986essays,valencia1991studies,maccartney2009extend,icard2014recent}. Similar to the monotone functions in calculus, in natural language upward monotone keeps the entailment relation when the argument ``increases'' (e.g., \textit{some cats are playing} $\sqsubset$ \textit{some animals are playing}, where \textit{cats} is replaced by its hypernym \textit{animals}). Downward monotone keeps the entailment relation when the argument ``decreases'' (e.g., \textit{all animals are playing} $\sqsubset$ \textit{all cats are playing}, where \textit{animals} is replaced by its hyponym \textit{cats}). 

To extend the monotonicity to consider exclusion, \newcite{maccartney2009extend} investigate all sixteen equivalence classes of \textit{set relations} and remove nine degenerate, semantically vacuous relations, thereby defining a seven-relation set $\frak{B} = \{\,\equiv, \sqsubset, \sqsupset, \wedge, \,\mid\,, \smallsmile, \#\,\}$ for natural logic, as shown in Table~\ref{table:basic-rel}.



From a high-level perspective, the natural logic proof system proposed by~\newcite{maccartney2009extend} consists of the following steps. First, the alignment between two text spans (often two sentences) is obtained and then lexical relation recognition  is performed for aligned pairs of words. Consider a simplified example: a premise \textit{All animals outside are eating} and a corresponding hypothesis \textit{All cats outside are playing}, as shown in Figure~\ref{fig:nnl_structure}. Each pair of aligned words is assigned one of the relations in Table~\ref{table:basic-rel}, e.g., \textit{animals} $\sqsupset$ \textit{cats} and \textit{eating} $\mid$ \textit{playing}. 

Projection $\rho \colon \frak{B} \rightarrow \frak{B}$ is then performed according to the projectivity in specific context. The projection operation has been implemented in the Stanford \textit{natlog} parser\footnote{https://stanfordnlp.github.io/CoreNLP/natlog.html}. For a given sentence, \textit{natlog} can output the projections at each word position. For example, Table~\ref{table:all} summarizes the projections in the context of the quantifier  \textit{all}, \textit{some}, and \textit{no}.    
Specifically, consider the example we discussed in the last paragraph: as \textit{animals} and \textit{cats} take place in the first argument of the quantifier \textit{all},
according to the projectivity in Table~\ref{table:all}, the \textit{reverse entailment} relation (\textit{animals} $\sqsupset$ \textit{cats}) will be projected to \textit{forward entailment} (\textit{animals} $\sqsubset$ \textit{cats}) in this specific context. As another example, since \textit{eating} and \textit{playing} take place in the second argument of \textit{all}, the \textit{alternation} relation (\textit{eating} $\mid$ \textit{playing}) is projected to \textit{alternation} (\textit{eating} $\mid$ \textit{playing}). 

Built on this, relation aggregation is performed to aggregate multiple projected local relations, according to Table~\ref{table:union}, to determine the global relation between the sentence pair. In our example, two projected relations, \textit{forward entailment} ($\sqsubset$) and \textit{alternation} ($\,\mid\,$), are aggregated to yield \textit{alternation} ($\,\mid\,$); i.e., we obtain \textit{All animals outside are eating} $\mid$ \textit{All cats outside are playing}. The seven natural logic relationships at the sentence level can be used to determine NLI relations. For example, if NLI is defined as a three-way classification problem (\textit{entailment}, \textit{contradiction}, and \textit{neutral}). The `$\,\equiv\,$' or `$\,\sqsubset\,$' relation will be mapped to \textit{entailment}, the `$\,\wedge\,$' or `\,$\mid$\,' relation will be mapped to  \textit{contradiction}, and `$\,\sqsupset\,$', `$\,\smallsmile\,$', or `$\,\#\,$' to \textit{neutral}.

\section{Neural Natural Logic Model}
\begin{figure*}
\centering

\begin{minipage} [t]{1\linewidth}
    \includegraphics[width=\linewidth, trim={0 0 0 0},clip]{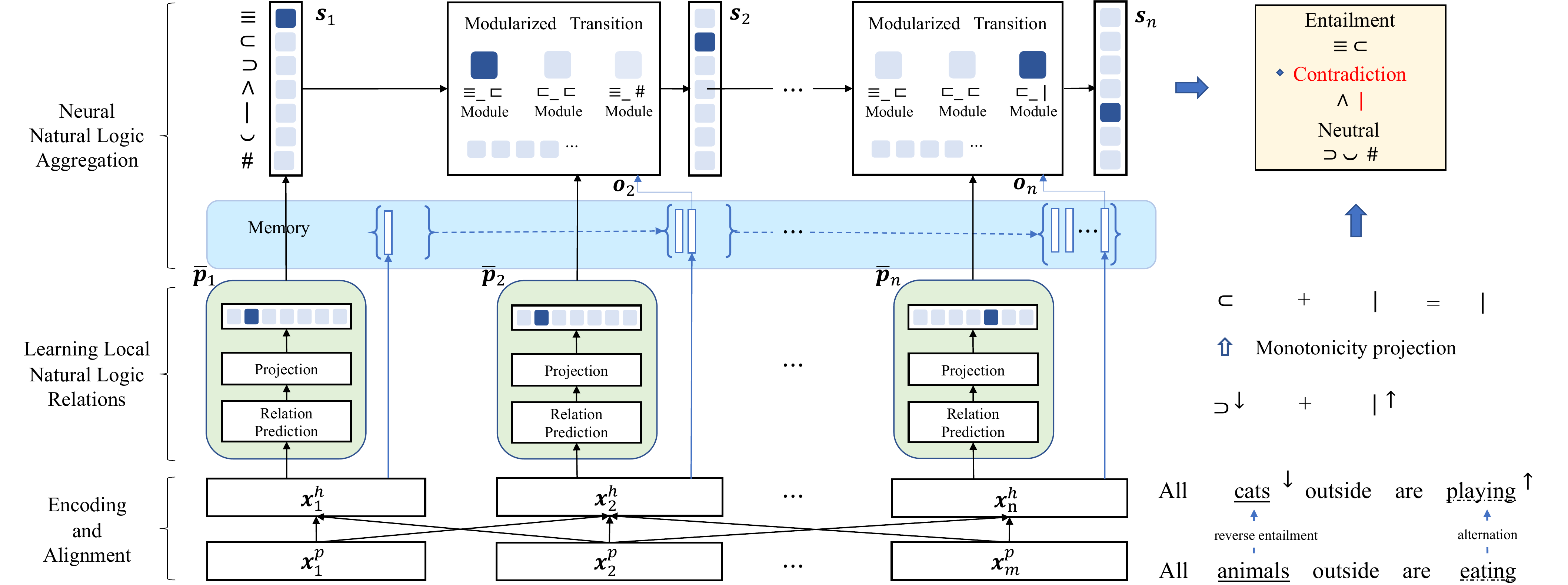}
\label{fig:nnlc_path}
\end{minipage}

\caption{A high-level view of the proposed neural natural logic model.}
 
\label{fig:nnl_structure}
\end{figure*}

We present a differentiable framework in which natural logic is integrated with neural networks. The overall architecture of the model is shown in Figure~\ref{fig:nnl_structure}. 
%
At the core of the framework are natural logic operations modeled with memory-enhanced module networks, which are trained end-to-end to optimize the following objective:
\begin{align}
    p(y|\boldsymbol{X}) = \sum_{\boldsymbol{z}\in \mathcal{Z}} p(y|\boldsymbol{z})p(\boldsymbol{z}|\boldsymbol{X})
\end{align}
where $y$ is the output, which in natural language inference is the label of the relation between a premise and hypothesis sentence (e.g.,\,\textit{entailment}, \textit{contradiction}, and \textit{neutral}), and which can be different labels in other tasks.
The input $\boldsymbol{X}=\langle \boldsymbol{X}^p,\boldsymbol{X}^h \rangle$  comprises a  premise sentence $\boldsymbol{X}^p$ and a hypothesis sentence $\boldsymbol{X}^h$. We use $\boldsymbol{z} = \{z_1, z_2, ..., z_n\}$ to denote a sequence of latent variables corresponding to the output of natural logic aggregation at each time step, where $n$ is the number of hidden variables.
The term $\mathcal{Z}$ denotes the space of all possible trajectories and $\boldsymbol{z} \in \mathcal{Z}$. Specifically, for the example in Figure~\ref{fig:nnl_structure}, if we perform the aggregation from left to right, 
$z_1=\,$`$\,\equiv\,$', $z_2=z_3=z_4=\,$`$\,\sqsubset\,$', and $z_5=\,$`$\,\mid\,$' is a $\boldsymbol{z}$ trajectory that proves the \textit{contradiction} label. Note that $z_i \in \frak{B}$ where $\frak{B}$ is the set of seven relations listed in Table~\ref{table:basic-rel}.

\subsection{Encoding and Alignment}
\label{sec:encoding_align}

Recent research has shown the effectiveness of distributed representations for encoding lexicons and their semantic relations. We use word embedding and neural networks to learn lexical representations to capture natural logic related semantics. Let $\boldsymbol{X}^p = \{\,\boldsymbol{x}_1^p, \boldsymbol{x}_2^p,...,\boldsymbol{x}_m^p\,\}$ be a premise sentence and  $\boldsymbol{X}^h =\{\,\boldsymbol{x}_1^h, \boldsymbol{x}_2^h,..., \boldsymbol{x}_n^h\,\}$ the corresponding hypothesis sentence, where $m$ and $n$ are the number of word tokens in the premise and hypothesis, repspectively. Each sentence is fed into a multi-layer BiLSTM, for which $\boldsymbol{a}_i = \bilstm(\boldsymbol{X}^p, i)$ denotes the $i^{\,\text{th}}$ hidden vector at the top layer of the BiLSTM, encoding the $i^{\,\text{th}}$ token and its context in the premise. Similarly, we use $\boldsymbol{b}_j = \bilstm(\boldsymbol{X}^h, j)$ to denote the hidden vector at the $j^{\,\text{th}}$ position at the top layer of the BiLSTM that encodes the hypothesis.
In this paper, we focus on understanding neural natural logic itself, without being further confounded by different ways of exploring knowledge external to the training data, e.g., via pretraining. 

Many models can be used to capture cross-sentence attention. Focusing on the training data, the approach proposed in \cite{esim2017} has been widely used in the NLI literature as a baseline. We follow  the work to compute cross-sentence attention weight $e_{ij} = \boldsymbol{a}_i^T\boldsymbol{b}_j \label{eqn:eij}$ for each pair $\langle\boldsymbol{a}_i, \boldsymbol{b}_j\rangle$. Specifically, for each $\boldsymbol{b}_j$ in the hypothesis, the corresponding content in a premise is weighted summed as $\Tilde{\boldsymbol{b}}_j = \sum_{i=1}^m\frac{\exp(e_{ij})}{\sum_{k=1}^m \exp(e_{kj})}\boldsymbol{a}_i
$, which will be used together with $\boldsymbol{b}_j$ to learn local lexical-level inference relations (refer to \cite{esim2017} for more details). 

In addition, we compute a \textit{hard alignment indicator} $\phi_j$, and $\phi_j = 1$ \textit{if and only if} 
$\boldsymbol{x}^p_{i^*} = \boldsymbol{x}^h_j$, where $i^* = \argmax_{i\in \{1, ..., m\}} e
'_{ij}$.\footnote{Here $e'_{ij}$ is the cross-attention weight obtained from the ESIM model ~\cite{esim2017} trained on SNLI.} That is, for each word token $\boldsymbol{x}^h_j$ in the hypothesis, we record the token $\boldsymbol{x}^p_{i^*}$ in the premise that has the maximum attention value $e'_{ij}$. If the word token $\boldsymbol{x}^p_{i^*}$ and $\boldsymbol{x}^h_j$ are the same word type, we let $\phi_j = 1$, which will be used to help reduce the search space in aggregation.

\subsection{Learning Local Natural Logic Relation} 
\label{sec:local_inference}

Given a sequence of alignment \{\,$\langle\Tilde{\boldsymbol{b}}_1, \boldsymbol{b}_1\rangle, ..., \langle\Tilde{\boldsymbol{b}}_j, \boldsymbol{b}_j\rangle, ..., \langle\Tilde{\boldsymbol{b}}_n, \boldsymbol{b}_n\rangle$\,\}, we use a bi-linear model to compute each pair's probabilistic distribution $\boldsymbol{p}_j$ over the natural logic relations $\frak{B}$:
\begin{align}
    \boldsymbol{p}_j = \softmax(f_s(\Tilde{\boldsymbol{b}}_j, \boldsymbol{b}_j)) = \softmax(\Tilde{\boldsymbol{b}}_j^T \mathcal{M}^T \boldsymbol{b}_j)
\end{align}
In the scoring function $f_s$, each type of relation $k \in \frak{B}$ has its own weight matrix $\mathcal{M}_k \in \mathbbm{R}^{d \times d}$, which is a slice of the tensor $\mathcal{M} \in \mathbbm{R}^{d \times d \times |\frak{B}|}$, where $d$ is the dimensionality of $\boldsymbol{b}_j$ or $\Tilde{\boldsymbol{b}}_j$. We use \textit{softmax} to normalize the values to be a
distribution over $\frak{B}$. Among several alternatives we used, the bi-linear model achieves the best performance on the development dataset, and we use it in our final framework.

\subsubsection{Local Relation Constraints}
Same as in many other weakly supervised setups, we do not have direct supervision signals here to learn logic relationships at the lexical level; instead, the supervision signals are backpropagated from the overall sentence-level NLI errors. To reduce the search space and alleviate the \textit{spurious} problem~\cite{guu2017language}
 in which incorrect local inference relationships and aggregation produce correct sentence-level NLI labels,\footnote{In an extreme case, if a model predicts the first aligned word pair between a premise and hypothesis to be a relation that is consistent with the ground-truth NLI label at the sentence level, the model can choose to ignore all other pairs that follow, and make the correct sentence-level prediction by using the first pair prediction only, even if the aggregation sequence $\boldsymbol{z}$ is incorrect.} we adopt several strategies as follows.

\label{sec:constraints}
\paragraph{Symmetric Inference Parameter Sharing:}  
We make the \textit{forward entailment}~($\sqsubset$) and \textit{reverse entailment}~($\sqsupset$) relations share the same parameters. Specifically, to compute $p^{\sqsupset}_j$, we reverse the order of $\langle\Tilde{\boldsymbol{b}}_j, \boldsymbol{b}_j\rangle$ to reuse  $\mathcal{M}_{\sqsubset}^T$ in the following scoring function, where $\mathcal{M}_{\sqsubset}^T$ is a matrix in $\mathcal{M}^T$ that corresponds to the  \textit{forward entailment}~($\sqsubset$) relation.
\begin{align}
f_s^{\sqsupset}(\Tilde{\boldsymbol{b}}_j, \boldsymbol{b}_j) = f_s^{\sqsubset}(\boldsymbol{b}_j, \Tilde{\boldsymbol{b}}_j) = \boldsymbol{b}_j^T \mathcal{M}_{\sqsubset}^T \Tilde{\boldsymbol{b}}_j
\end{align}


\paragraph{Equivalence Constraint:} A token pair will be assigned the \textit{equivalence} relation~($\equiv$), if $\phi_j$ learned above in the alignment stage takes the value of 1:  
    \begin{equation}
    \text{if  }\phi_j = 1, \text{ we let\hphantom{z}} p^{\equiv}_j = 1\quad   
    \end{equation}

\paragraph{Collapse Constraints:} We suppress the relations \textit{negation}~($\wedge$) and \textit{cover}~($\smallsmile$):
    \begin{equation}
        p^{\wedge}_j = 0, \quad p^{\smallsmile}_j = 0
    \end{equation}

Inspired by \newcite{angeli2014naturalli}, we suppress the \textit{negation} relation~($\wedge$)  because its behavior is almost same as that of \textit{alternation}~($\,\mid\,$) in natural logic aggregation, as shown in Table \ref{table:union}, avoiding the co-linearity problem when training on datasets without double negation samples. We also suppress the \textit{cover} relation ($\smallsmile$) because it is extremely rare in current natural language inference datasets.

\subsubsection{Projected Distribution}
\label{sec:mono_projection}
With the predicted seven-dimensional probability vector $\boldsymbol{p}_j$ being ready, our model uses a projection operator $\rho$ to re-organize the distribution according to the projectivity of the corresponding input hypothesis word at position $j$. 
%
%
Unlike the discrete ``hard" projection used in the conventional natural logic, e.g.,\,projecting the first argument of \textit{all} from \textit{reverse entailment} to \textit{forward entailment}, we apply ``soft" projection over relation probability distribution $\boldsymbol{\bar{p}}_j$. Specifically, based on the projection Table~\ref{table:all}, we convert the original probability distribution $\boldsymbol{p}_j$ to the projected distribution $\boldsymbol{\bar{p}}_j$:
\begin{align}
    \bar{p}^{k'}_j = \sum_{k} p^k_j \mathbbm{1}(\rho(k) = k') \label{eqn:mono_project},
\end{align}
where $\mathbbm{1}(\cdot)$ is the indicator function, $k$ is the original relation, and $k'$ is the projected relation. Consider the pair of sentences in Figure~\ref{fig:nnl_structure} and suppose the pair \textit{eating} vs. \textit{playing} have a probability of 0.8 to be \textit{alternation}~(\,$\mid$\,) and 0.1 to be \textit{negation}~($\wedge$). According to the projectivity of the second argument of the quantifier \textit{all} in Table~\ref{table:all}, both relations are projected to alternation (\,$\mid$\,): $\rho^{playing}(\,\mid\,) = \rho^{playing}(\wedge) = \,\mid$. 
So after projection, $\bar{p}^{|}_{5} = p^{|}_{5} + p^{\wedge}_{5} = 0.9$, where the subscript $5$ is the index of the word token \textit{playing} in the hypothesis.

 
\subsection{Aggregation}
\label{sec:nnl_reasoning}


\begin{figure}[t]
\centering

\begin{minipage} [t]{1\linewidth}
\centering
    \includegraphics[width=0.9\linewidth,clip]{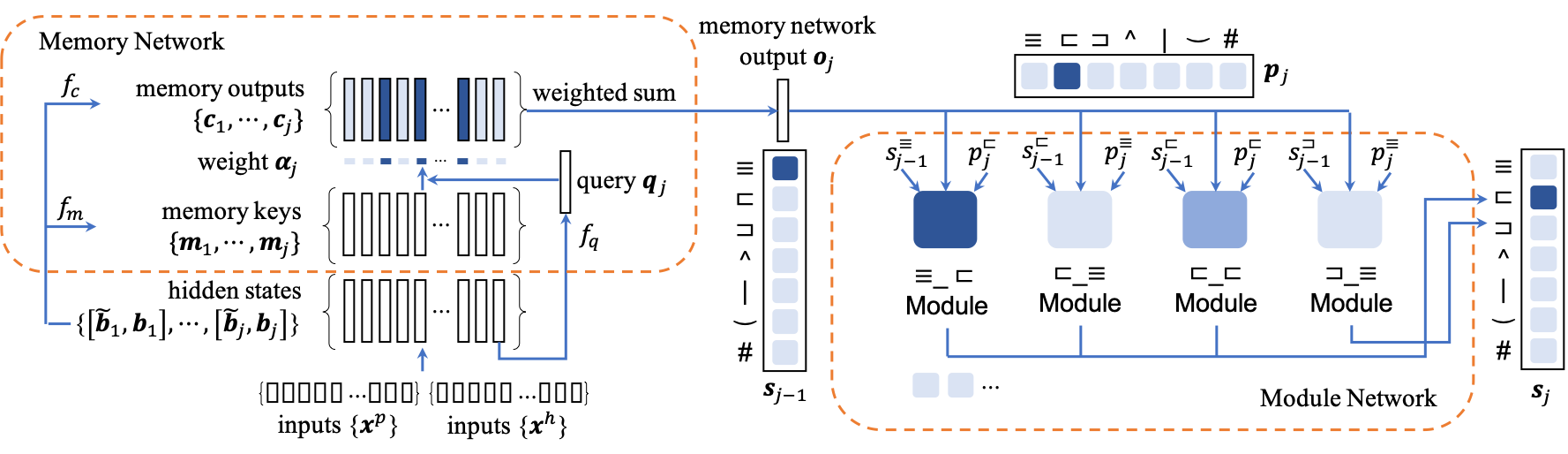}
\end{minipage}

\caption{A memory-enhanced module network for natural logic aggregation.}
 
\label{fig:module}
\end{figure}



We propose to leverage the module networks \cite{Andreas_2016_CVPR,gupta2019neural} to perform neural natural logic aggregation, which is enhanced by a memory network component to leverage the powerful ability in modeling context. Figure \ref{fig:module} shows the proposed neural natural logic aggregation network. The right part of the figure is the aggregation module network and the left is the memory network component.


Specifically, at each time step $j$, our aggregation algorithm computes a distribution $p(z_{j}|\boldsymbol{X}) = \softmax(\boldsymbol{s}_{j})$, where $\boldsymbol{s}_{j} = \{s^k_{j}\}$ is a set of logits. $s^k_{j}$ is the one corresponding to $p(z_{j}=k|\boldsymbol{X})$ for relation $k \in \frak{B}$. Our model computes $s^k_{j}$ with Equation~\ref{eqn:aggregation}.

\begin{align}
    s^k_{j} =& \sum_{u \in \frak{B}}\sum_{v \in \frak{B}} G^{u \Join v}(s^u_{j-1}, \bar{p}^v_{j}, \boldsymbol{o}_{j}) \mathbbm{1}(u \Join v = k) \nonumber\\ \label{eqn:aggregation}
            =& \sum_{u \in \frak{B}}\sum_{v \in \frak{B}}[s^u_{j-1} \cdot \bar{p}^v_{j}  \cdot g^{u \Join v}(\boldsymbol{o}_{j})] \mathbbm{1}(u \Join v = k) \nonumber \\
            =& \sum_{u \in \frak{B}} s^u_{j-1} \sum_{v \in \frak{B}} \bar{p}^v_{j}  \cdot g^{u \Join v}(\boldsymbol{o}_{j}) \mathbbm{1}(u \Join v = k),
\end{align}

At time step 1, $\boldsymbol{s}_{1}$ is initialized with $\bar{\boldsymbol{p}}_{1}$. At any other time step $t>1$, we invoke modules $G^{u \Join v}(\cdot)$ to derive $\boldsymbol{s}_{j}$. Specifically, in our network each relation aggregation in Table~\ref{table:union}, i.e., $u \Join v$ ($u, v \in \frak{B}$), has its own module $G^{u \Join v}(\cdot)$. Now, given the previous $\boldsymbol{s}_{j-1}$ and the current \textit{projected local relation} distribution $\bar{\boldsymbol{p}}_{j}$,  $\boldsymbol{s}_{j}$ can be computed by marginalizing the Cartesian product $\boldsymbol{s}_{j-1}  \cdot \boldsymbol{\bar{p}}^T_{j} $ according to aggregation Table~\ref{table:union}. More specifically, we first compute the Cartesian product $\boldsymbol{s}_{j-1}  \cdot \boldsymbol{\bar{p}}^T_{j} $, which is weighted by the memory $g^{u \Join v}(\boldsymbol{o}_{j})$. Then for all modules with output being the same 
relation $k$ according to Table~\ref{table:union}, the modules' output are summed up, where $\mathbbm{1}(\cdot)$ is the indicator function.

Below we discuss how the memory network response $\boldsymbol{o}_{j}$ is calculated.
In this paper, we propose a memory network component \cite{weston2014memory,sukhbaatar2015end} to enhance our module aggregation network, aiming to better model contextual information. The details are shown in the left part of Figure~\ref{fig:module}. 
Specifically, at time step $j$, we store memory vectors $\{\,\boldsymbol{m}_1,...,\boldsymbol{m}_j\,\}$ and the corresponding output vectors $\{\,\boldsymbol{c}_1,...,\boldsymbol{c}_j\,\}$ in the memory. The query vector $\boldsymbol{q}_j$ scans the memory and computes the match between itself and memory vectors by taking the inner product followed by a \textit{softmax}:
\begin{align}
    \boldsymbol{q}_j &= f_q([\boldsymbol{\Tilde{b}}_j, \boldsymbol{b}_j]) \\
    \boldsymbol{m}_j &= f_m([\boldsymbol{\Tilde{b}}_j, \boldsymbol{b}_j]) \\
    \boldsymbol{c}_j &= f_c([\boldsymbol{\Tilde{b}}_j, \boldsymbol{b}_j]) \\
    \alpha_{
    j,t} = \softmax &(\boldsymbol{q}_j^T\boldsymbol{m}_t), \ t=1,...,j
\end{align}

The query, memory, and output vectors are functions of aligned token representation $[\Tilde{\boldsymbol{b}}_j, \boldsymbol{b}_j]$, typically modeled by two feed-forward layers.
The response vector $\boldsymbol{o}_j$ is computed by the weighted sum over stored outputs vectors $\boldsymbol{c}_j$  and is used in the module network discussed above:  
\begin{align}
    \boldsymbol{o_j} = \sum_{t=1}^j \alpha_{j, t} c_t
\end{align}
where $\boldsymbol{o}_j$ encodes all historical transitions and their context and is then incorporated into Equation~\ref{eqn:aggregation}.

In addition to the sequential aggregation we discuss above in which we perform aggregation left-to-right over a premise and hypothesis pair, we also perform the aggregation on the binarized constituency parses, where aggregation is performed on a tree structure. For node $j$ in the constituency tree, we define a random variable $z_j$ which represents the reasoning states upon seeing the node $j$ and sub-tree, and we use $\boldsymbol{s}_j$ to denote the distribution of $z_j$. We initialize $\boldsymbol{s}_j$ with projected relation distribution $\bar{\boldsymbol{p}}_j$ if node $j$ is the leaf node. Iteratively, the distribution $\boldsymbol{s}_j$ for each non-leaf node is computed by aggregating its left child ($lc$) and right child ($rc$): 
\begin{align}
    s^k_{j} &= \sum_{u \in \frak{B}}\sum_{v \in \frak{B}} G^{u \Join v}(s^u_{lc}, s^v_{rc}, \boldsymbol{o}_{j}) \mathbbm{1}(u \Join v = k)
\end{align}
where $\boldsymbol{o}_j$ is the memory network response vector which is computed on the information of all nodes that have already been visited. 


\paragraph{Objective Function}
The final prediction of sentence relation is computed with the distribution of hidden state $\boldsymbol{s}_n$ at the last time step (or the root node if reasoning is performed over the constituency tree). We follow the work of \newcite{angeli2014naturalli} and group relation \textit{equivalence}~($\equiv$) and \textit{forward entailment}~($\sqsubset$) to be \textit{entailment};  \textit{negation}~($\wedge$) and \textit{alternation}~(\,$\mid$\,) to be \textit{contradiction}, and; \textit{reverse entailment}~($\sqsupset$), \textit{cover}~($\smallsmile$) and \textit{independent} ($\#$) to be \textit{neutral}. We apply a variant of hard-EM training method~\cite{min2019discrete}, which selects the most likely relation: $p_{entailment} = \max(s^{\equiv}_n, s^{\sqsubset}_n)$,  $p_{contradiction} = \max(s^{\wedge}_n, s^{\mid}_n)$, and $p_{neutral} = \max(s^{\sqsupset}_n, s^{\smallsmile}_n, s^{\#}_n)$. 
After applying softmax, we obtain the prediction probability, which can be used to compute the cross entropy loss.

\section{Experiments}
\subsection{Setup}

\paragraph{Data:}
We use three datasets that are designed for studying monotonicity based reasoning, i.e., HELP~\cite{help2019}, MED~\cite{med2019}, and the monotonicity subset of Semantic Fragments~\cite{fragments2020}. 
The HELP dataset has 35,891 inference pairs, which are automatically generated by conducting lexical substitution or deletion on one sentence to obtain the other, given natural logic polarity information of each word token and syntactic structure of sentences.
The MED dataset contains 5,382 human-generated inference pairs by either asking crowdworkers to perform the generation or manually collecting the pairs from linguistics publications. The monotonicity subset of Semantic Fragments is automatically generated with a controlled set of rules and lexicons, which contains around 2,000 pairs. Since the pairs with the contradiction relation in the Semantic Fragments dataset are obtained by changing quantifiers, which are out of the scope of the natural logic formalism that we use, we do not include this subset in our experiments. 

In addition, we create a new \textit{2-hop} dataset. The above datasets lack ground-truth labels for evaluating aggregation at each time step, and most of them are \textit{1-hop} aggregation in which a premise and hypothesis differs only by one span of text. In our \textit{2-hop} dataset, the premise and hypothesis differs by two edits of word/phrase insertion, deletion, or substitution. Our dataset provides ground-truth aggregation output \{\,$z_1$, ..., $z_j$, ... $z_n$\,\} to help assess models' performance on natural logic operations and understand their decision paths. The development of this 2-hop dataset includes three steps: (a) identify the editing type for each example in MED and determine the logic relations; (b) add one more \textit{hop} of relation, and; (c) record the ground-truth aggregation labels at each time step and the final NLI labels following MacCartney's natural logic formalism. We manually checked a subset of the data and found more than 96\% of examples are correct. Details of 
the data development are included in Appendix \ref{appendix:2hop}.

 
\paragraph{Implementation Details:}
Following \newcite{esim2017}, hidden vectors in our model are 300 dimensional. We use pretrained 300-dimensional 840B GloVe vectors~\cite{glove2014} to initialize our word embeddings. All word embeddings are trainable after being initialized. We apply a dropout  rate of $p=0.5$. Adam~\cite{kingma2014adam} is used as our optimizer, and the ﬁrst momentum is set to be 0.9 and the second 0.999. The batch size is set to be 32 and the initial learning rate is 0.0004. We train ESIM and our neural natural logic models for 32 epochs and use the development set to select models for testing. We use default hyper-parameters specified in \cite{bert2019} and train the BERT-base model for 3 epochs.


\subsection{Results}
\begin{table*}
\setlength{\tabcolsep}{1.6pt}\small
\centering
\begin{tabular}{c|cccc|cc}\toprule
 \multirow{2}{*}{\textbf{Model}} &    \textbf{Monotonicity} &    \multirow{2}{*}{\textbf{HELP} (\%)}   &   \multirow{2}{*}{\textbf{MED} (\%)} &   \textbf{Natural Logic} &\textbf{HELP Dev} (\%)  & \textbf{HELP Test} (\%)\\
  &     \textbf{Fragments} (\%)  &    & & \textbf{2-Hop} (\%)& Up Mono.& Down Mono.\\
\midrule
 \textbf{ESIM}&     66.18   &   55.27   &   51.78   & 45.13 & 95.25 & 21.49\\
 \textbf{BERT-base}  &      50.58   &   51.40   &  45.88   &   49.33& 98.63 & 13.71\\
\midrule

\textbf{Neural Nat. Log. (seq.)}  &      66.03   &   58.23   &   \textbf{52.47}  &  \textbf{60.14} &91.20 & 63.08\\
\textbf{Neural Nat. Log. (tree)}  &     \textbf{66.47}   &   \textbf{63.95}   &   47.57  &  59.97 & 90.62 & \textbf{70.80} \\
\bottomrule
\end{tabular}
\caption{Test accuracy of the models.}\label{table:main_acc}

\end{table*}


 
\paragraph{Inference Performance:}
Table \ref{table:main_acc} shows the test accuracy of different models on the four datasets that are designed specifically for evaluating monotonicity-based inference. Following \newcite{fragments2020} and \newcite{med2019}, we train the models on SNLI \cite{snli2015} and test on these different test sets. The proposed models, in general, achieve better performances on these four datasets than ESIM \cite{esim2017} and BERT \cite{bert2019}. The difference is more prominent in the 2-hop dataset, which requires the system to have a better aggregation ability to make the final prediction. 

To demonstrate how the models generalize between the upward and downward monotone, we train the models with HELP's upward monotone subset and test on the downward monotone subset. A system that can better model monotonicity should achieve more robust performance. Specifically, we split the upward monotone subset of the HELP dataset into the training set ($\sim6$k training examples) and the development set ($\sim1.5$k examples). We train all models on the training split and select models with the highest development accuracy. We test all models on the HELP downward monotone subset ($\sim21$k examples). The right-most column of Table~\ref{table:main_acc} shows that while ESIM and BERT achieve very high development accuracy on the upward data, they fail to generalize to the downward monotone test set. The proposed models generalize well and achieve better test accuracy on the downward monotone datasets. 

\begin{table*}
\setlength{\tabcolsep}{3pt}\small
\centering
\begin{tabular}{c|l|ccc}\toprule
&\multicolumn{1}{c|}{\textbf{Model}} & \textbf{Precision} &    \textbf{Recall} &    \textbf{F1} \\
\midrule

(1)& \textbf{Neural Nat. Log. (seq.)}  &\textbf{0.54} &   \textbf{0.49} & \textbf{0.51}  \\
(2)&(1)  w/o. \textbf{Memory \vphantom{-} / \vphantom{-} Module} & 0.49 & 0.46 & 0.47 \\
(3)&(2)  w/o. \textbf{Local Rel. Constraints} &  0.12 &   0.15 & 0.13 \\
\bottomrule
\end{tabular}
\caption{Evaluation of models' aggregation performance on the 2-hop dataset.}
\label{tab:explain}
\end{table*}

\begin{figure*}
    \centering
       \includegraphics[width=0.95\linewidth, trim={1.4cm 1cm 1.2cm 0cm},clip]{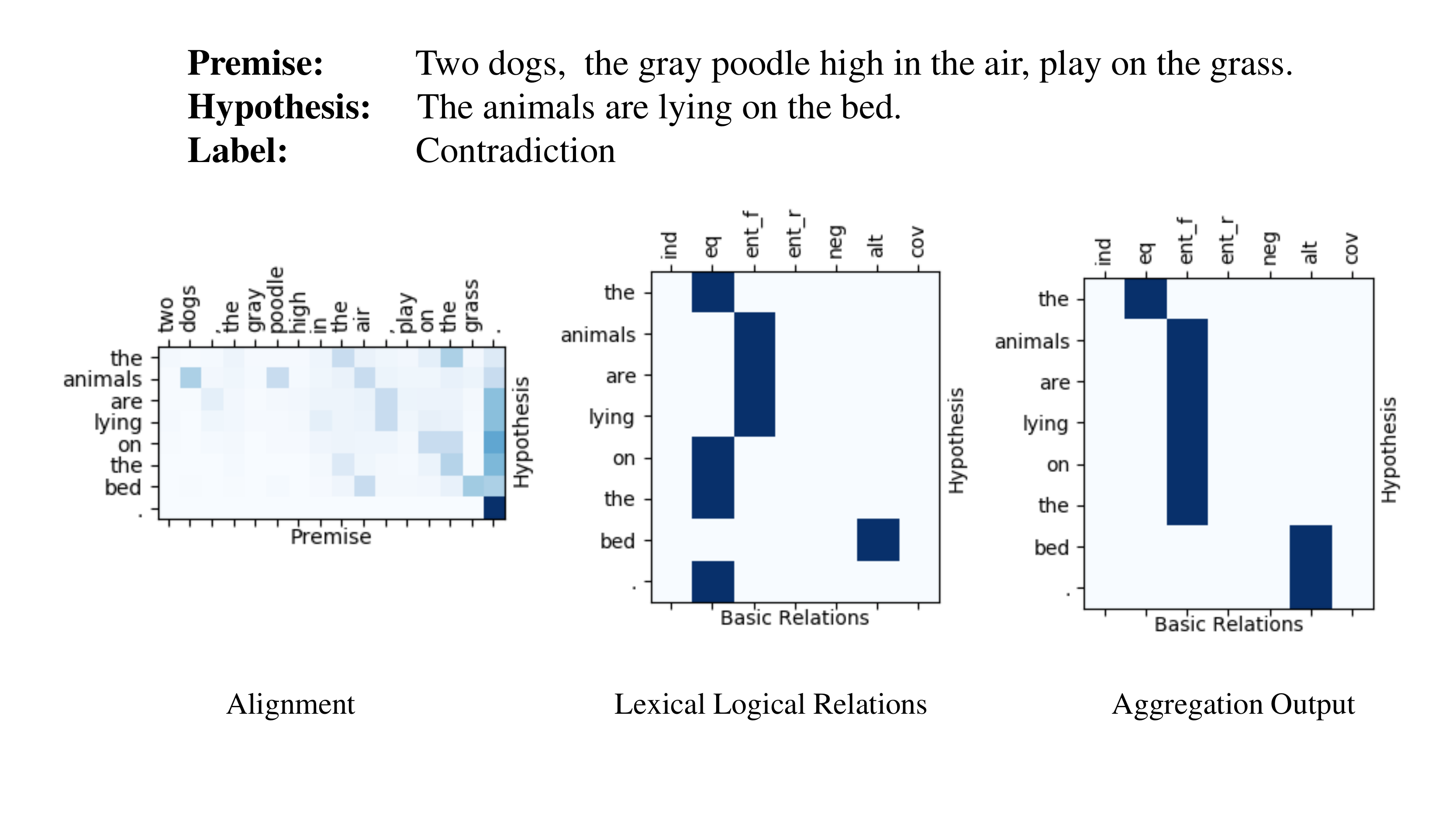}
    \caption{An example showing how the proposed model perform natural logic aggregation.}
    \label{fig:case}
\end{figure*}

\paragraph{Aggregation Decisions:} 
The proposed model provides inference explainability by accessing natural logic's aggregation and decision paths.
Figure~\ref{fig:case} shows an example of the 2-hop dataset, together with the visualization of the intermediate aggregation decisions. From left to right, the first subfigure shows the cross-sentence attention between the premise (x-axis) and hypothesis (y-axis), where a darker color corresponds to a larger attention weight. In the second subfigure, for each word in the hypothesis (y-axis), the predicted distribution of lexical-level logical relations are shown along the x-axis. The third subfigure shows the aggregation output. For example,
on the second row, the aggregation has already been performed over the first two words $\boldsymbol{b}_1$~=\,``\textit{the}" and $\boldsymbol{b}_2$ =\,``\textit{animals}" using their lexical relation distributions, which have been shown in the second subfigure and are, in turn, computed from the first subfigure using $\langle\boldsymbol{\Tilde{b}}_1, \boldsymbol{b}_1\rangle$ and $\langle\boldsymbol{\Tilde{b}}_2, \boldsymbol{b}_2\rangle$. Since `$\equiv$' $\Join$ `$ \sqsubset$' \,=\, `$\sqsubset$', we can see that on the second row, a large probability mass has been put on $\sqsubset$ (i.e., \textit{ent\_f} in the figure).  


We further perform quantitative analysis on the aggregation performance. We analyze the sequential aggregation. Specifically, for the 2-hop dataset in which we have access to the aggregation decisions: $\hat{\boldsymbol{z}} = \{\hat{z}_1, \hat{z}_2, ...,\hat{z}_n\}$, where $\hat{z}_j$ is the aggregation result at time step $j$, we evaluate the models by comparing the estimated $\hat{\boldsymbol{z}}$ with the ground truth $\boldsymbol{z}$. We use precision, recall, and F-score as our evaluation metrics. The details of how to compute them are in Appendix~\ref{appendix:metrics}.

Table~\ref{tab:explain} shows the results. Since ESIM and BERT do not produce intermediate aggregation results, they are not included in the table. The ablation analysis shows that both the memory/module component and the local relation constraints help the model to learn intermediate natural logic aggregation. We can also see that further work is desirable to improve the performance on aggregation prediction as there is still a large room to improve modeling performance on this. As part of our efforts, we have also performed component training to leverage WordNet \cite{wordnet} and ConceptNet \cite{conceptnet} to help determine lexical relations. This approach is not particularly effective since the lexical pairs from these knowledge bases only cover a very small percentage of pairs that need to be modeled.

\section{Conclusions}



This paper studies end-to-end trained differentiable models that integrate natural logic with neural networks. The proposed model adapts module networks to model natural logic operations, which is enhanced with a memory component to model contextual information. We analyze the proposed model on the monotonicity subset of Semantic Fragments, HELP, MED, and a subset of MED that are modified to include 2-hop inference. 
Our experiments show that the proposed framework can effectively model monotonicity-based reasoning, compared to the two baseline neural network models without built-in inductive bias for monotonicity-based reasoning. The proposed model show to be robust when transferred from upward to downward inference. We perform further analyses on the performance of the proposed model on aggregation, showing the effectiveness of the proposed subcomponents on helping achieve better intermediate aggregation performance.

\section{Acknowledgement}
The first, second, fourth and last author's research is supported by NSERC Discovery Grants.

\bibliographystyle{coling}
\bibliography{coling2020}



\newpage
\appendix
\section{The 2-hop Dataset}
%
\label{appendix:2hop}
Figure~\ref{fig:datasample} shows an example of the 2-hop dataset. The premise and hypothesis differ by two edits of word/phrase insertion, deletion, or substitution. The dataset provides the ground truth of aggregation at each time step (the \textit{equivalence} relation is the default relation and is hence not included in the ``Ground truth of aggregation" section) and the word locations/indices associated with each edit. The 2-hop dataset is developed with the following three steps:
\begin{figure}[H]
    \centering
    \includegraphics[width=0.83\linewidth, trim={5cm 4cm 4.5cm 4cm},clip]{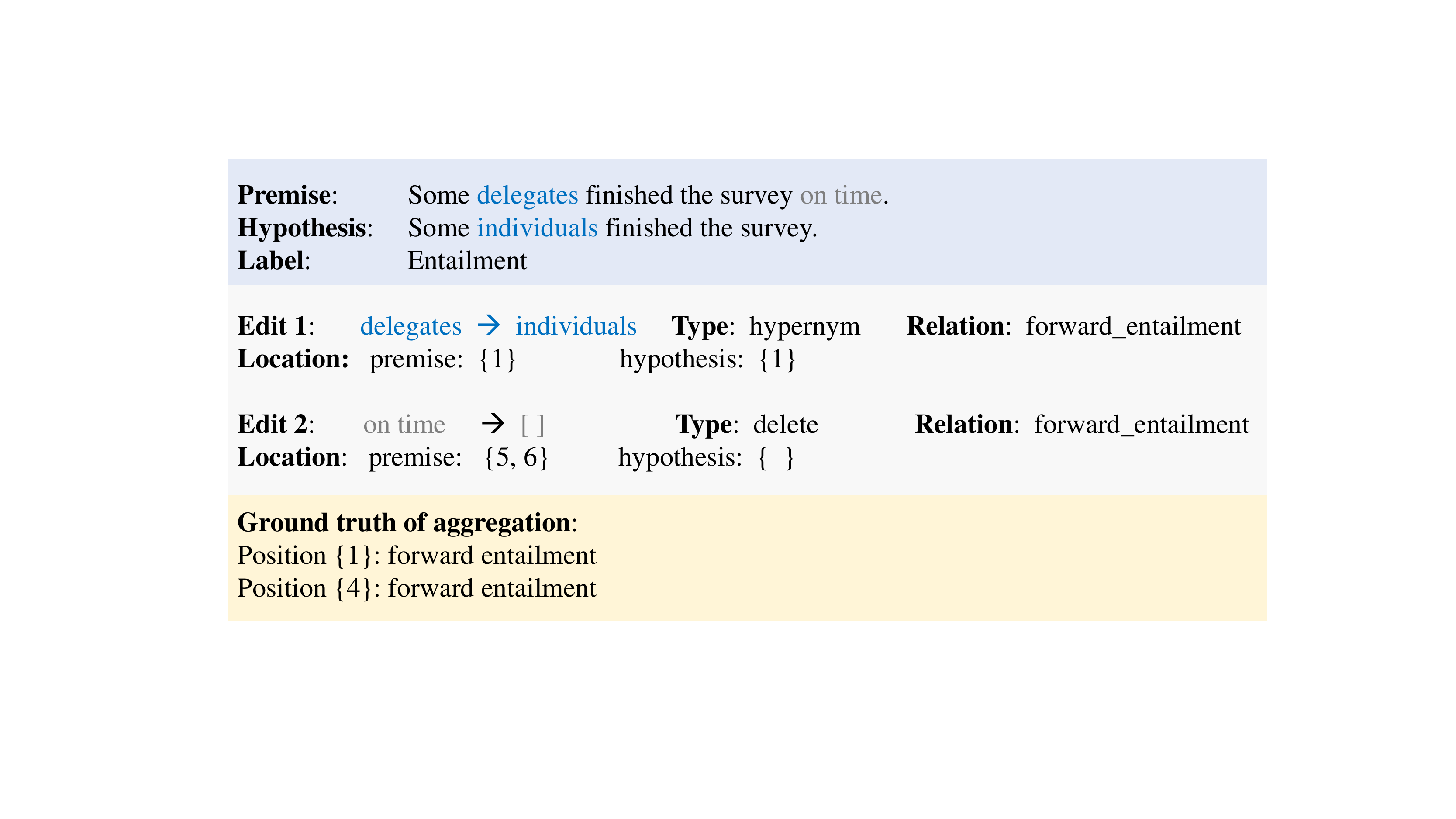}
    \vspace{-15 pt}
    \caption{An example of the 2-hop dataset.}
    \label{fig:datasample}
\end{figure}
\paragraph{Identifying MED Relations:} 
Since most sentence pairs in the MED dataset are only different by one word/phrase edit; i.e.,\,the premise and the hypothesis differs by one word/phrase, it is easy to determine location of the insertion, deletion, or replacement. For insertion and deletion, we follow \cite{angeli2014naturalli} and treat the relation as \textit{reverse entailment} ($\sqsupset$) and \textit{forward entailment} ($\sqsubset$), respectively. We set aside the replacement samples since we can not determine their relations without human labeling. To ensure the identified natural logic relations are correct, we compare the labels provided in MED with labels determined by MacCartney's natural logic theory and remove samples in which labels do not agree, yielding roughly 1.1K sentence pairs.

\paragraph{Adding One More Hop of Relations:} 
We ask human annotators to replace a noun either in the premise or the hypothesis with another word. The relation between the substituted and substituting word are one of $\{\equiv, \sqsubset, \sqsupset, \mid, \# \}$. Annotators have access to WordNet that can help suggest substituting words (e.g., hypernyms or hyponyms). Meanwhile, we require that the candidate words to be replaced are not  children or parents of any previously identified differences over the parsing tree. This replacement operation yields 5,858 sentence pairs, and the premise and the hypothesis of each example now differ by two edits.

\paragraph{Determining Labels:} 
We apply projection operation and natural logic aggregation according to \cite{maccartney2009extend} to determine the 3-way natural language inference labels for the generated 2-hop sentence pairs. We also record the ground-truth relations of each hop of aggregation output.   
We manually assess the data quality on 300 sentence pairs (100 for each category). We find that on average 3\% of the samples have either incorrect labels or wrong intermediate aggregation output (4\% in category \textit{Entailment}, 4\% in category \textit{Neutral} and 1\% in \textit{Contradiction}). Those mistakes are mainly produced by incorrect parser-identified polarity.


\section{Aggregation Evaluation Metrics}
\label{appendix:metrics}
We evaluate the intermediate aggregations of the proposed model with the precision, recall, and F1 score.
Precision is the number of correctly performed aggregations, divided by the total number of aggregations performed by a model. Recall is the number of correctly performed aggregations, divided by the total number of aggregations presented in the ground-truth annotation. Note that we only consider aggregations at time step $t$ when $\hat{z}_t \neq \hat{z}_{t-1}$. Since by default the starting state $\hat{z}_{0} = \,$`$\equiv$', so if $\hat{z}_{1} = \,$`$\equiv$', we do not count this degenerate case.


\end{document}